\documentclass[10pt,twocolumn,letterpaper]{article}

\usepackage{cvpr}              % To produce the CAMERA-READY version
\usepackage{graphicx}
\usepackage{amsmath}
\usepackage{amssymb}

\usepackage{tabularx}
\usepackage{adjustbox}
\usepackage{times}
\usepackage{booktabs}
\usepackage{microtype}
\usepackage{epsfig}
\usepackage[table,xcdraw]{xcolor}
\usepackage{caption}
\usepackage{float}
\usepackage{placeins}
\usepackage{color, colortbl}
\usepackage{stfloats}
\usepackage{enumitem}
\usepackage{tabularx}
\usepackage{xstring}
\usepackage{xspace}
\usepackage{url}
\usepackage{subcaption}
\usepackage{multirow}
\usepackage{multicol}
\usepackage{xcolor}
\usepackage[hang,flushmargin]{footmisc}

\usepackage[normalem]{ulem}
\useunder{\uline}{\ul}{}
\usepackage{booktabs}
\usepackage[switch]{lineno}

\usepackage[pagebackref,breaklinks,colorlinks]{hyperref}

\usepackage[capitalize]{cleveref}
\crefname{section}{Sec.}{Secs.}
\Crefname{section}{Section}{Sections}
\Crefname{table}{Table}{Tables}
\crefname{table}{Tab.}{Tabs.}

\def\cvprPaperID{9372} % *** Enter the CVPR Paper ID here
\def\confName{CVPR}
\def\confYear{2023}

\begin{document}

\title{Seeing What You Miss: Vision-Language Pre-training with \\ Semantic Completion Learning}

% % just like the second author
% \and
\def\authorBlock{
    Yatai Ji$^\textnormal{1}$\footnotemark[1] \qquad
    Rongcheng Tu$^\textnormal{2}$\footnotemark[1] \qquad
    Jie Jiang$^\textnormal{2}$\footnotemark[1] \qquad
    Weijie Kong$^\textnormal{2}$ \qquad
    Chengfei Cai$^\textnormal{2}$ \\
    Wenzhe Zhao$^\textnormal{2}$ \qquad
    Hongfa Wang$^\textnormal{2}$ \qquad
    Yujiu Yang$^\textnormal{1}$\footnotemark[2] \qquad
    Wei Liu$^\textnormal{2}$\footnotemark[2]
    \and
    $^\textnormal{1}$Tsinghua University \qquad
    $^\textnormal{2}$Tencent \\
    {\tt\small jyt21@mails.tsinghua.edu.cn} \qquad {\tt\small turongcheng@gmail.com} \\
    {\tt\small \{zeus, jacobkong, fletchercai, carsonzhao, hongfawang\}@tencent.com } \\
    {\tt\small yang.yujiu@sz.tsinghua.edu.cn} \quad {\tt\small wl2223@columbia.edu}
}
\author{\authorBlock}
\maketitle

{
  \renewcommand{\thefootnote}%
    {\fnsymbol{footnote}}
  \footnotetext[1]{Equal contribution.}
  \footnotetext[2]{Corresponding Author.}
}

%%%%%%%%% ABSTRACT
\begin{abstract}
\vspace{-0.2cm}
% Vision-language pre-training (VLP) models have attracted much research attention due to their remarkable performance on many downstream vision-language tasks. 
Cross-modal alignment is essential for vision-language pre-training (VLP) models to learn the correct corresponding information across different modalities. 
For this purpose, inspired by the success of masked language modeling (MLM) tasks in the NLP pre-training area, numerous masked modeling tasks have been proposed for VLP to further promote cross-modal interactions. 
% in the fusion encoders. 
The core idea of previous masked modeling tasks is to focus on reconstructing the masked tokens based on visible context for learning local-to-local alignment. 
However, most of them pay little attention to the global semantic features generated for the masked data, resulting in a limited cross-modal alignment ability of global representations. 
% which has a great impact on the performance of pre-trained models on downstream tasks. 
% Therefore, in this paper, we propose a novel \textbf{S}emantic \textbf{C}omplementation \textbf{L}earning (SCL) task, complementary to existing masked modeling tasks, promoting to learn more representative global features. 
Therefore, in this paper, we propose a novel \textbf{S}emantic \textbf{C}ompletion \textbf{L}earning (SCL) task, complementary to existing masked modeling tasks, to facilitate global-to-local alignment. 
% global semantic representations to learn cross-modal alignment. 
Specifically, the SCL task complements the missing semantics of masked data by capturing the corresponding information from the other modality, promoting learning more representative global features which have a great impact on the performance of downstream tasks. 
Moreover, we present a flexible vision encoder, which enables our model to perform image-text and video-text multimodal tasks simultaneously. 
Experimental results show that our proposed method obtains state-of-the-art performance on various vision-language benchmarks, such as visual question answering, image-text retrieval, and video-text retrieval.
\end{abstract}
\vspace{-0.6cm}

%%%%%%%%% BODY TEXT
\section{Introduction}

\begin{figure}[tp]
\centering
  \includegraphics[width=0.49\textwidth]{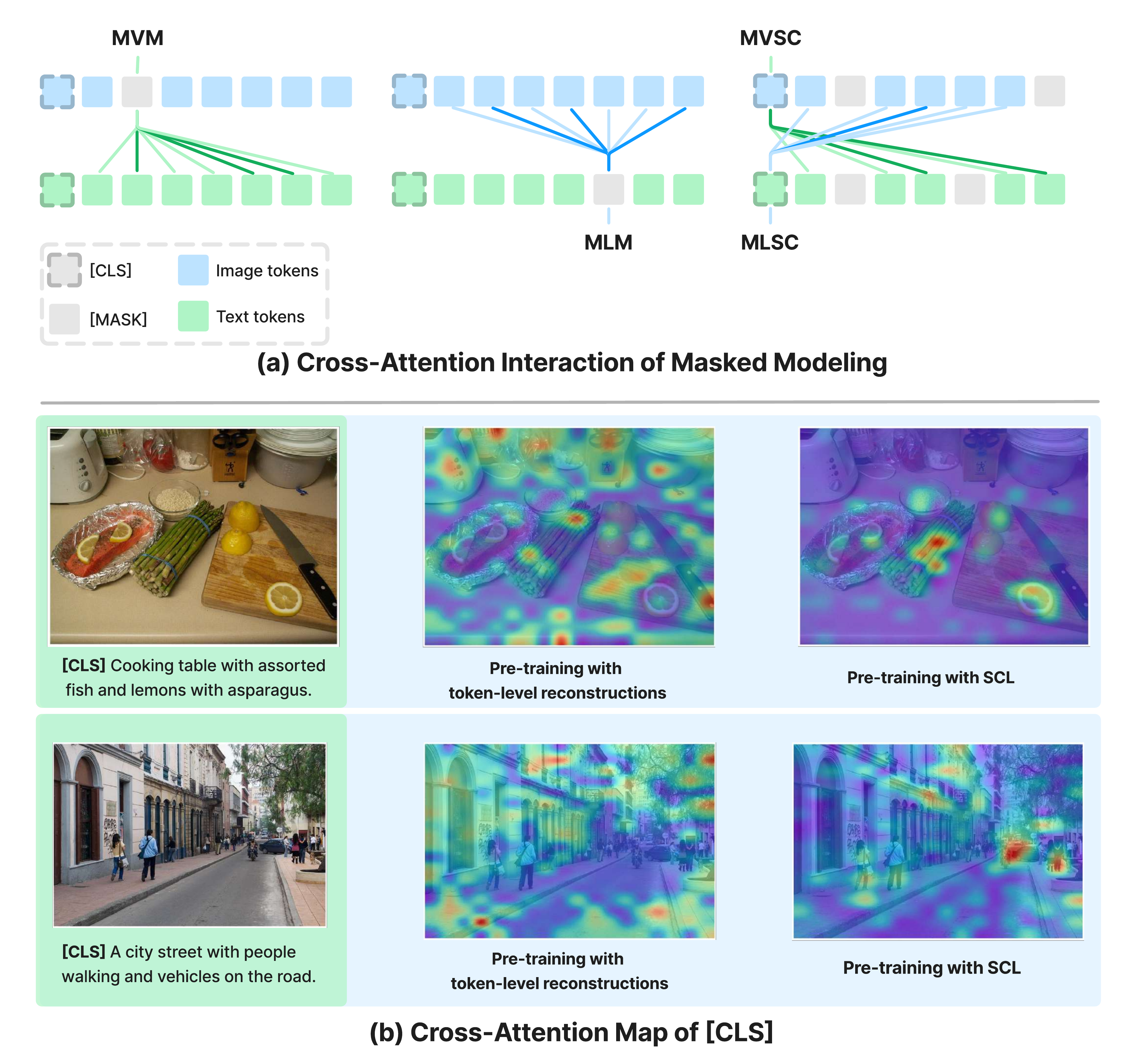}
  \vspace{-0.3cm}
  \caption{(a) The comparisons between previous masked modeling tasks and our proposed Semantic Completion Learning (SCL), which is composed of ``MVSC'' and ``MLSC''. (b) The cross-modal attention map visualization of the text global representation ([CLS]) on the input image for our model pre-trained with or without SCL.}
  \label{fig:intro_visulize}
  \vspace{-0.5cm}
\end{figure}

Our real-world contains a wide variety of information, such as texts, images, sounds, etc. 
% A powerful general artificial intelligence system should understand multimodal semantic information like human, learning knowledge from different modality sources. 
For a powerful general artificial intelligence system, it is necessary to capture the semantic association from different modality sources.
%There is semantic association from different modality sources, which is necessary for a powerful general artificial intelligence system to understand. 
Towards this goal, multimodal representation learning is a crucial technique to bridge the heterogeneity gap between different modalities~\cite{DBLP:journals/corr/abs-2109-04290/dualsoft, DBLP:journals/corr/abs-2206-08916/unifiedio}. 
% Pre-training has impressive performance in individual modalities, such as computer vision and natural language processing. 
% 单模态预训练模型
% To obtain the common sense of semantic alignment, It is essential for multimodal learning to pre-train from huge multimodal data, which can significantly improve downstream tasks performance, for instance, visual question answering and image-text retrieval. 
In this area, vision-language pre-training models \cite{DBLP:conf/emnlp/TanB19/lxmert,DBLP:conf/nips/LuBPL19/vilbert,DBLP:conf/acl/LiGNXLL0020/unimo,DBLP:conf/cvpr/GeGLLSQL22/mcq,DBLP:journals/corr/abs-2210-05335/map} have shown an impressive semantic alignment ability,  which brings substantial advances on various downstream tasks, for instance, visual question answering, image-text retrieval, etc.
% Mask-then-predict is a mainstream self-supervised manner in multimodal pre-training. 
% In masked language modeling, the model leverages vision and text context to rebuild masked tokens. 
% In terms of vision modality, the input of current models is patch sequence for efficiency. 
% Some works regress raw pixel values of masked patches following MAE, which is proven to advance retrieval performance.\kwj{Why?} 
% Some works predict discrete codes of patches as BeiT does, but has to generate annotations with dVAE first. 
% After masking out input images or text, the above tasks only recover the fine-grained content based on the corresponding local information from the other modality. 
% They are beneficial for fine-grained cross-modal alignment but lack explicit reconstruction of global semantics. \kwj{The motivation is not so attractive!}
% When the input content is disturbed, the high-level semantic information is also corrupted. \kwj{Not always.}
% It may lead to a suboptimal global representation if only the reconstruction of local content is considered. 
% Furthermore, downstream tasks such as retrieval and VQA directly rely on global representations. 
% Then, a natural question comes in: since the modeling of local information can't guarantee the recovery of high-level semantics, can we recover the global representations of masked data? 

Recently, numerous self-supervised vision-language pre-training models~\cite{DBLP:conf/emnlp/TanB19/lxmert, DBLP:conf/eccv/Li0LZHZWH0WCG20/oscar, fu2021violet, DBLP:conf/cvpr/GeGLLSQL22/mcq, bain2021frozen,DBLP:journals/corr/abs-2208-09374/vlmae, DBLP:journals/corr/abs-2206-01670/egocentric} have been proposed. 
These methods model the interactions between vision and language features mainly by using various masked modeling tasks, such as masked language modeling (MLM) and masked vision modeling (MVM). 
As shown in Fig.~\ref{fig:intro_visulize}(a), the basic idea of MLM and MVM is self-reconstructing the masked tokens via leveraging informative visible tokens to realize local-to-local alignment. 
Specifically, MLM adopted by BERT~\cite{kenton2019bert} is to predict the original vocabulary IDs of the masked words. 
Inspired by the success of MLM in pre-training, there is a flourishing trend to extend it to visual pre-training tasks. 
Generally, by masking some visual patches, MVM tasks predict their original pixels~\cite{DBLP:journals/corr/abs-2208-09374/vlmae, DBLP:conf/mm/Gao0LGWLY22/calic}, corresponding discrete tokens~\cite{DBLP:journals/corr/abs-2206-01127/vlbeit, DBLP:journals/corr/abs-2208-10442/beit3, fu2021violet} generated by the VQ-VAE variants, or Histograms of Oriented Gradients (HOG) features~\cite{DBLP:journals/corr/abs-2209-01540/violet2}, etc.

% It can be seen these masked modeling tasks only focus on reconstructing the local masked tokens, and pay little attention to recover the missed global semantic information caused by data corruption. 

% Although these masked modeling tasks proposed by previous works are beneficial to improve the performance of vision-language pre-training models, they only focus on reconstructing the local masked tokens. 

These masked modeling tasks only focus on reconstructing the local masked tokens, and pay little attention to recovering the missing global semantic information caused by data corruption. 
The token-level reconstruction may lead to inadequate learning of global representations for cross-modal information. 
%As illustrated in Fig~\ref{fig:intro_visulize}(b), in the situation of token-level reconstructions, the global representation is disordered in its attention on the information of the other modality, meaning the lack of global-to-local alignment ability. 
% As illustrated in Fig~\ref{fig:intro_visulize}(b), in the situation of token-level reconstructions, the global representation is disordered in its attention on the information of the other modality.
As illustrated in Fig.~\ref{fig:intro_visulize}(b), in the situation of token-level reconstructions, the global representation is disordered in its attention on the other modality. 
It implies that the global-to-local alignment ability of the pre-training model is limited, leading to a degraded global representation. 
However, the global semantic features have a great impact on the performance of the pre-training model as they are usually used to deal with downstream tasks. 
Therefore, it is crucial to ensure the global semantic features to learn more accurate global-to-local alignment. 
% Considering that the paired vision and text data are two views of the same semantic information, intuitively, the global semantics of masked data can be recovered by capturing information from their corresponding data of the other modality. 

Intuitively, considering that the paired vision and text data are two views of the same semantic information, the missing semantics of masked data can be completed by capturing information from the other modality. 
From this point of view, we propose a novel pre-training task called \textbf{S}emantic \textbf{C}ompletion \textbf{L}earning (SCL). 
Specifically, SCL is composed of dual parts: masked vision semantic completion (MVSC) and masked language semantic completion (MLSC). 
%The input of vision or language modality is masked respectively and fused with complete data from the other modality. 
As shown in Fig.~\ref{fig:intro_visulize}(a), MVSC (MLSC) exploits information of complete text (vision) data to recover the global semantic representations of masked vision (text) data. 
% In this way, the model can complete the missing semantics and learn more accurate cross-modal alignment for global features. 
In this way, the model can generate representative global features with accurate global-to-local alignment. 
For example, as illustrated in Fig.~\ref{fig:intro_visulize}(b), compared with the model pre-trained without SCL, the attention maps with SCL pre-training are more discriminative and reasonable. 

For the architecture of the vision-language pre-training model, we adopt a general framework that consists of two uni-modal encoders and a fusion encoder. 
Moreover, we present a flexible vision encoder to enable our model to perform image-text and video-text multimodal tasks simultaneously. Specifically, for video inputs, the vision encoder only adds a few additional learning parameters, and the [CLS] feature of each frame is treated as a bridge associating spatial modeling within the frame and temporal modeling among frames. 
Inspired by curriculum learning~\cite{bain2021frozen}, we train the model with image-text and video-text datasets successively to transfer visual knowledge from images to videos.

In a nutshell, our contributions are three-fold. 
(1) To enhance the global-to-local alignment of global representations, we propose a new pre-training task called Semantic Completion Learning (SCL), which recovers missing semantic information from unmasked data, promoting learning more representative global features. 
(2) We design an adaptive vision encoder, which can transfer multimodal pre-training knowledge between images and videos readily. 
% Intuitively, the vision encoder contributes to making image-text and video-text pre-training in synergy. 
(3) We conduct multiple vision-language downstream tasks to demonstrate the generalization of semantic completion learning, and the vision encoder, including visual question answering, visual reasoning, image-text retrieval, and video-text retrieval. 
Our model SCL achieves state-of-the-art performance based on a similar pre-training data scale.
Our code is available at https://github.com/IIGROUP/SCL. 
%the pre-training data of similar size. 
% Detailed ablation studies depict the effect of different pretext tasks. 

\section{Related Works}
\subsection{Vision-Language Pre-training}
Existing vision-language pre-training works can be divided into two categories: dual-tower and cross-fusion architecture.  

The dual-tower architecture based methods \cite{DBLP:conf/icml/RadfordKHRGASAM21/clip,andonian2022robust,jia2021scaling,gabeur2020multi,xu2021videoclip,bain2021frozen,akbari2021vatt} employ two individual encoders to separately extract the features for the visual data (images or videos) and textual data, and then map these features into a common semantic space. Among them, CLIP \cite{DBLP:conf/icml/RadfordKHRGASAM21/clip} exploits contrastive learning with a huge quantity of noisy image-text pairs directly collected from the Internet, achieving remarkable results on plenty of vision-language tasks. 
Similarly, FROZEN \cite{bain2021frozen} proposes a curriculum learning schedule to train the vision-language model on both image-text and video-text datasets by treating an image as a single-frame video. 
% and the experimental results demonstrate the effectiveness of the curriculum learning schedule.
Although these two-stream architecture based methods perform well on cross-modal retrieval tasks with high efficiency, their performances on the more complex multimodal downstream tasks are not inspirational due to the insufficient interaction between local vision and text features.
% Similarly, VATT \cite{akbari2021vatt} employs multimodal contrastive learning to align the video frames, audios and texts, and achieves impressive performance on the downstream tasks. FROZEN \cite{bain2021frozen} proposes a curriculum learning schedule to train the vision-language model on both image-text and video-text datasets by treating an image as a single-frame video, and the experimental results demonstrate the effectiveness of the curriculum learning schedule. Although these two-stream architecture based methods perform well on cross-modal retrieval tasks with high efficiency, their performances on the more complex multimodal downstream tasks are not inspirational due to the insufficient interaction between local visual-texture features.

To overcome this limitation, the cross-fusion architecture based methods~\cite{DBLP:conf/emnlp/TanB19/lxmert, DBLP:conf/acl/LiGNXLL0020/unimo, chen2020uniter, DBLP:conf/eccv/Li0LZHZWH0WCG20/oscar, Dou_2022_CVPR/meter} have been proposed, which employ a cross-modal fusion encoder to enhance the interactions between vision and text features. For example, ALBEF \cite{akbari2021vatt} not only aligns the image and text features with contrastive learning but also feeds them into a cross-modal attention-based encoder to obtain the fused features. 
Clover~\cite{DBLP:journals/corr/abs-2207-07885/clover} improves cross-modal feature alignemnt and fusion via a tri-modal alignment pre-training task. 
Our model also conducts multimodal feature fusion to achieve encouraging performance on more downstream tasks. 
% to calculate the matching scores. Moreover, VIOLET \cite{fu2021violet} proposes a masked visual-token modeling task to train a joint encoder for the vision-language fusion.

\subsection{Masked Modeling Tasks}
Recently, various masked modeling tasks have been proposed, whose strategy is self-reconstructing the masked data. Masked Language modeling (MLM) adopted by BERT \cite{kenton2019bert} is the most classical one. It randomly masks some tokens of the input and then predicts the original vocabulary IDs of the masked words based on their context. 
By pre-training with the MLM, BERT achieves state-of-the-art results on eleven natural language processing (NLP) tasks. 
Inspired by the success of MLM in NLP, some works extend it into the visual domain and propose masked vision modeling (MVM). 
% After masking some visual patches, some MVM works mainly predict the corresponding discrete tokens~\cite{DBLP:journals/corr/abs-2206-01127/vlbeit, DBLP:journals/corr/abs-2208-10442/beit3, fu2021violet} generated by VQ-VAE or its variants; other works directly regresses the original pixels~\cite{DBLP:journals/corr/abs-2208-09374/vlmae, DBLP:journals/corr/abs-2208-02131/maevl} or Histograms of Oriented Gradients (HOG) features~\cite{DBLP:journals/corr/abs-2209-01540/violet2} of masked visual tokens. 
For example, VLMAE~\cite{DBLP:journals/corr/abs-2208-09374/vlmae} proposes the Regional Masked Image Modeling (RMIM) task to facilitate the fusion of multimodal features. The RMIM masks some patches of an input image and then reconstructs the original pixels depending on the visible patches and the corresponding text. 
Similarly, VIOLET \cite{fu2021violet} proposes a masked visual-token modeling task, which first maps the original video frame patches into discrete visual tokens and then recovers the corresponding visual tokens of masked patches to train a joint encoder for the vision-language fusion.
% Moreover, the MLM and MVM tasks recently also have been used in vision-language pre-training model to enhance the cross-modal interactions of their fusion encoders. 
% Overall, masked tokens in MLM and MVM are reconstructed via leveraging corresponding tokens in the other modality, which we named local-to-local alignment. 
However, these tasks focus on reconstructing local masked tokens, ignoring the recovery of global semantic information of the masked data after cross-modal interactions. 
Hence, we propose a novel semantic completion learning (SCL) task.

\section{Approaches}

In this section, we first introduce our pre-training  objectives in Sec.~\ref{pre-training task}, and then describe the model architecture in Sec.~\ref{model architecture}. 
Please refer to~\cref{appendix:model} for the whole architecture figure and the details of previous pre-training tasks.

\begin{figure*}[tp]
\centering
  \includegraphics[width=0.9\textwidth]{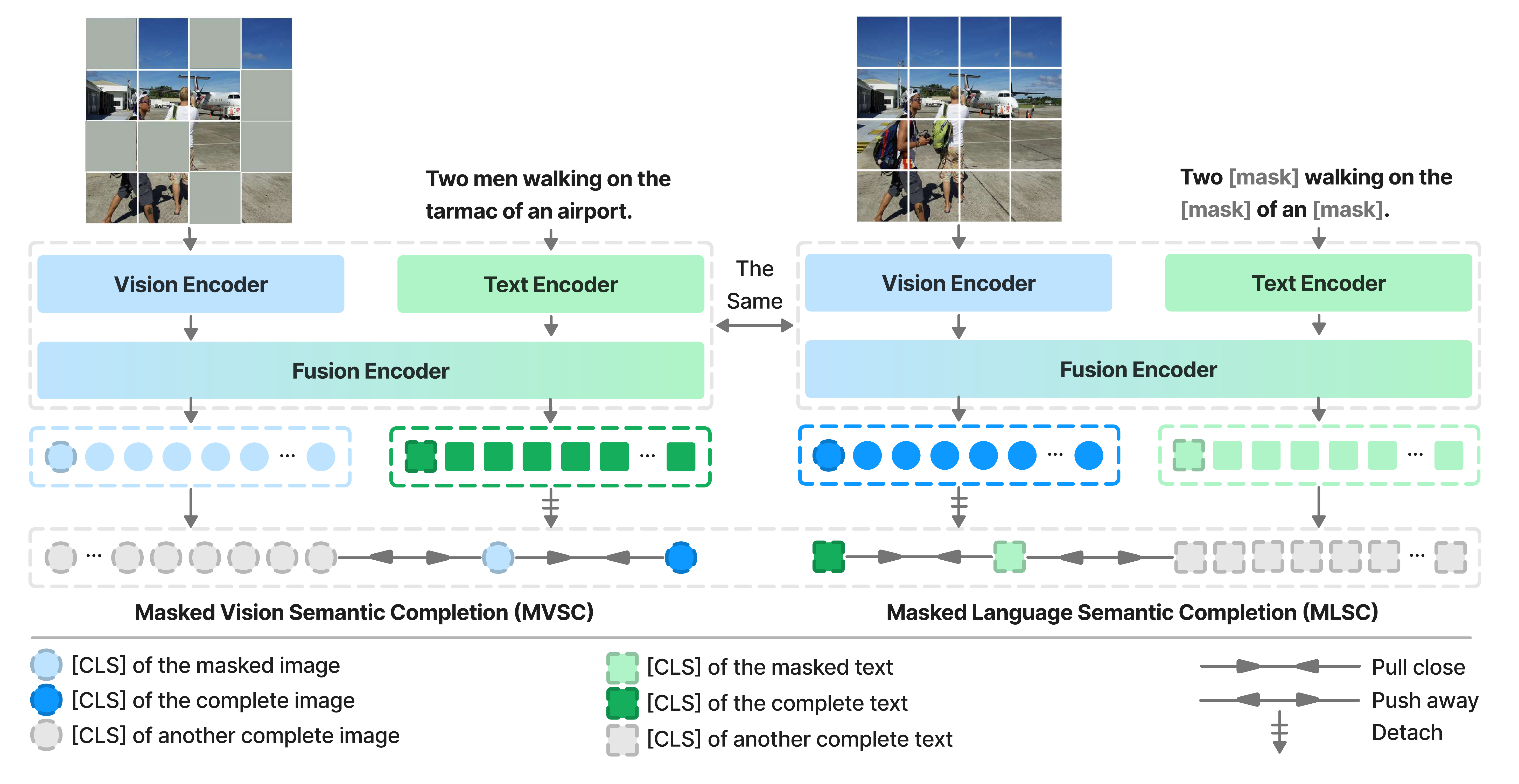}
  \vspace{-0.2cm}
  \caption{The overview of our proposed Semantic Completion Learning (SCL). The two versions of an image-text pair are forward propagated, respectively, to perform masked vision/language semantic completion.}
  \label{fig:SCL}
  \vspace{-0.4cm}
\end{figure*}

\subsection{Pre-training Tasks}
\label{pre-training task}
\subsubsection{Previous Pre-training Tasks}
% 之前的预训练任务除了常见的三个，剩下的复原像素对高级语义任务有害，复原vision token需要额外的编码器，或者视觉定位预训练任务需要区域和标注的额外监督。
\noindent\textbf{Contrastive Learning (CL).} 
%% ?
The input images and texts are projected into vision and language embedding spaces with two uni-modal encoders, respectively. 
We utilize contrastive learning to adjust the positions of semantic features, enforcing the paired image-text features close and negative samples far apart. 
Then the token-wise fusion is employed for features of different modalities in the unified semantic space. 

\noindent\textbf{Vision-Text Matching (VTM).} 
VTM aims to determine the correspondence of an image-text pair. 
The model conducts a binary classification on the concatenation of vision and text global representations generated by the fusion encoder, which contributes to the overall alignment of different modalities. 

\noindent\textbf{Masked Language Modeling (MLM).} 
MLM was first used as a pretext task in natural language processing and was later introduced to multi-modal pre-training. 
Following the text tokens masked out with a probability of 15\%, the model attempts to predict the original words based on visual information and textual context. 
The token-level reconstruction task plays an important role in the way that the model learns to associate linguistic words and visual entities, realizing local-to-local semantic alignment. 

\subsubsection{Semantic Completion Learning (SCL)}
It is significant for the model to learn multi-modal information fusion, that is, to extract knowledge from the other modality. 
Instead of local information reconstruction in former masked modeling tasks, we expect that the model can also recover the global semantics of masked images or texts after cross-modal interaction. 

As shown in Fig. \ref{fig:SCL}, for each data pair, we first randomly mask the image and text separately to get $\{I_{mask}, T\}$ and $\{I, T_{mask}\}$, so that the masked one manages to learn semantic information from the other complete modality.  
%We adapt a high mask ratio, 80$\%$ for images and 40$\%$ for text in details, in order to disrupt original content. 
Then the two couples of data are sent to the model respectively. 
The recovered features of masked data are obtained by leveraging information from the other modality to complete its missing semantic information: 
\begin{equation}
% \small
\begin{aligned}
I_{Re}, T_{Co}&=\operatorname{Model}(I_{mask}, T), \\
I_{Co}, T_{Re}&=\operatorname{Model}(I, T_{mask}), \\
\end{aligned}
\label{msm_forward}
\end{equation}
where $I_{Re}, T_{Re}$ are \textbf{re}covered global features of masked data, and $I_{Co}, T_{Co}$ refer to global features of \textbf{co}mplete data. 
Then we conduct masked vision semantic completion (MVSC) and masked language semantic completion (MLSC) simultaneously. 
Specifically, we bridge the gap between recovered global features and the complete ones in the form of contrastive learning. 
The InfoNCE loss is adopted to maximize the mutual information (MI) between two versions of the input data pair, $\{I_{mask}, T\}$ and $\{I, T_{mask}\}$: 
\begin{equation}
\begin{aligned}
\operatorname{\textbf{NCE}}_V=-\frac{1}{N}\sum_{i=1}^{N}\operatorname{log}\frac{\operatorname{exp}(s(I_{Re}^i,I_{Co}^i)/\tau)}{\sum_{n=1}^N\operatorname{exp}(s(I_{Re}^i,I_{Co}^n)/\tau)}\, , \\
\operatorname{\textbf{NCE}}_L=-\frac{1}{N}\sum_{i=1}^{N}\operatorname{log}\frac{\operatorname{exp}(s(T_{Re}^i,T_{Co}^i)/\tau)}{\sum_{n=1}^N\operatorname{exp}(s(T_{Re}^i,T_{Co}^n)/\tau)}\, ,
\end{aligned}
\label{eq:nceloss}
\end{equation}
where $s$ denotes cosine similarity and $\tau$ serves as the temperature hyper-parameter. 
The negative samples are global features of other complete images or texts in a batch. 
Note that $I_{Co}$ and $T_{Co}$ are detached for gradient backward, which makes the model more focused on the learning of recovering global features. 
Finally, the SCL loss is defined as: 
\begin{equation}
\mathcal{L}_{SCL}=\operatorname{\textbf{NCE}}_V+\operatorname{\textbf{NCE}}_L.
\label{eq:scl}
\end{equation}

By minimizing Eq.(\ref{eq:scl}), it will make the global feature $I_{Re}$ of the masked image similar to $I_{Co}$ of the complete image ($T_{Re}$ similar to $T_{Co}$). 
% In other words, the semantic information of masked data is recovered by cross-modal learning. 
To recover the semantic information of masked data, the global representations will learn supplementary knowledge from the corresponding tokens of the other modality, i.e., accurate global-to-local alignment. 
% The global features will be recorvered by aggregated the informative tokens from the other modal data
%% 加总结例子

\subsection{Model Architecture}
\label{model architecture}
Our model consists of three components: Vision Encoder, Text Encoder, and Fusion Encoder. 
% In the following, we will introduce  each component in detail.

\subsubsection{Vision Encoder}
\noindent\textbf{Input. }The vision encoder takes visual data (a video or image) $I\in \mathcal{R}^{M\times3\times H \times W}$ containing $M$ frame(s) of resolution $H \times W$ as input, and when $I$ is an image, $M=1$. The visual data $I$ is first split into $M \times N$ patches $x\in \mathcal{R}^{M\times 3\times N \times P \times P}$, where $P \times P$ is the size of patches and $N=HW/P^2$. Then, the patches $x$ are transformed into $M$ sequences of vision tokens $\boldsymbol{V}=\{\boldsymbol{v}_i\}_{i=1}^M \in \mathcal R^{M\times N \times D}$, where $\boldsymbol{v}_i \in  \mathcal R^{N \times D}$ denotes the sequence of tokens for the $i^{th}$ frame in the visual data and $D$ denotes the dimension of vision tokens. Next, a learnable [CLS] token is concatenated to every token sequence $\boldsymbol{v}_i$, and we obtain $\boldsymbol{V}=\{{\boldsymbol{v}_i}\}_{i=1}^M \in \mathcal R^{M\times (N+1) \times D}$. Finally, the tokens $\boldsymbol{V}$ are summed with learnable spatial positional embeddings  $\boldsymbol{E}^s \in \mathcal{R}^{(N+1) \times D}$ and temporal positional embeddings $\boldsymbol{E}^t \in \mathcal{R}^{M \times D}$:
\begin{equation}
	\begin{aligned}
		\boldsymbol{g}^0_{ij} = \boldsymbol{v}_{ij} + \boldsymbol{E}^t_i + \boldsymbol{E}^s_j,
	\end{aligned}
	\label{f1}
\end{equation}
where all patches in the same spatial location of different frames are given the same spatial positional embedding $\boldsymbol{E}^s_j$, and all patches in the same frame share the same temporal positional embedding $\boldsymbol{E}^t_i$. 
\begin{figure}[]
	\centering
	\includegraphics[width=0.8\linewidth]{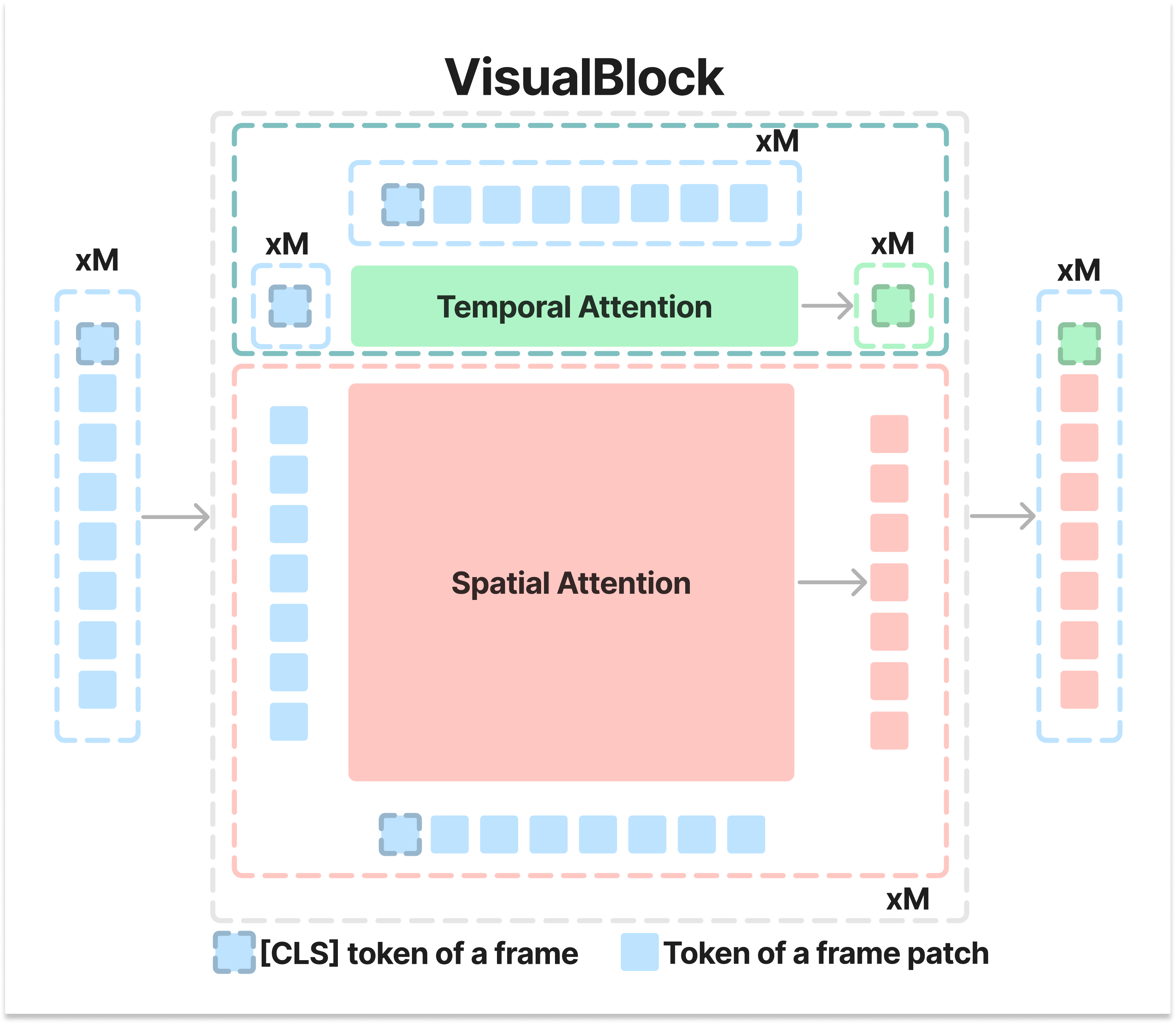}
	\vspace{-0.2cm}
	\caption{The architecture of a VisualBlock.}
	\label{fig_visual_encoder}
 \vspace{-0.4cm}
\end{figure}

\noindent\textbf{VisualBlock. }
The pre-processed vision tokens $\boldsymbol{G}^{0}=\{{\boldsymbol{g}^0_i}\}_{i=1}^M \in \mathcal R^{M\times (N+1) \times D}$ are fed into the vision encoder which can process image and video data. The vision encoder is a modified ViT~\cite{DBLP:conf/iclr/DosovitskiyB0WZ21/vit}, containing a stack of VisualBlocks. 
%Specifically, as shown in Fig. \ref{fig_visual_encoder}, we perform temporal and spatial attention, respectively, on the input $\boldsymbol{G}^{l-1}=\{{\boldsymbol{g}^{l-1}_i}\}_{i=1}^M$ of the $l^{th}$ VisualBlock. 

The detail of each VisualBlock is shown in Fig. \ref{fig_visual_encoder}.
Specifically, each VisualBlock will perform temporal attention to exploit the global temporal information of the visual data, and conduct the spatial attention to capture sufficient local spatial semantic information. 
For temporal attention, we perform multi-head attention for the [CLS] tokens $\{\boldsymbol{g}_{i0}^{l-1}\}_{i=1}^M$ of all frames through attending to all $M\times (N+1)$ tokens to produce [CLS] tokens $\{\boldsymbol{g}_{i0}^{l}\}_{i=1}^M$. 
Regarding spatial attention, it is the multi-head attention within each frame. 
Taking the $i^{th}$ frame as an example, we use $\{\boldsymbol{g}_{ij} ^{l-1}\}_{j=1}^N$ without [CLS] token as queries and all the $N+1$ tokens in the frame as keys and values to conduct attention and obtain the output tokens $\{\boldsymbol{g}_{ij} ^{l}\}_{j=1}^N$. 
After the temporal attention and spatial attention are conducted, we concatenate the $M$ [CLS] tokens $\{\boldsymbol{g}_{i0}^{l}\}_{i=1}^M$ with the $M \times N$ tokens $\{\{\boldsymbol{g}_{ij} ^{l}\}_{j=1}^N\}_{i=1}^M$ of frame patches as the output of the VisualBlock, denoted as $\boldsymbol{G}^{l}=\{{\boldsymbol{g}^{l}_i}\}_{i=1}^M$. 

\vspace{-0.2cm}
\subsubsection{Text Encoder}
\vspace{-0.2cm}
Given the input text $T$, we first tokenize it into word embeddings $\{\boldsymbol{t}_i\}_{i=1}^K$, where $K$ is the total number of words. Then, the text encoder, which consists of a stack of bidirectional transformers \cite{DBLP:conf/nips/VaswaniSPUJGKP17/transformer},
maps $\{\boldsymbol{t}_i\}_{i=1}^K$ into token features $\boldsymbol{W}=\{\boldsymbol{w}_i\}_{i=1}^K$ by modeling the contextual relationships.

\vspace{-0.2cm}
\subsubsection{Fusion Encoder}
\vspace{-0.2cm}
Similar to \cite{Dou_2022_CVPR/meter,DBLP:conf/emnlp/TanB19/lxmert}, we adopt a two-stream architecture for the fusion encoder, each layer of which consists of two modality-specific self-attention blocks and two cross-attention blocks. Specifically, taking the vision features as an example, following intra-modal interactions in the visual self-attention block, we conduct cross-modal interactions in the language-to-vision cross-attention block, which takes the vision tokens $\boldsymbol{G}=\{{\boldsymbol{g}_i}\}_{i=1}^M$ as queries and the text tokens $\boldsymbol{W}=\{\boldsymbol{w}_i\}_{i=1}^K$ as keys and values. The text features are conducted with similar operations. In the end, we use the mean pooling of all the frame [CLS] tokens yielded by the fusion encoder as the global representation for visual data and the [CLS] token of text as the global representation for text data.

%for text to vision fusing, we use the vision tokens $\boldsymbol{G}=\{{\boldsymbol{g}_i}\}_1^M$ as queries, the text tokens $\boldsymbol{W}=\{\boldsymbol{w}_i\}_{i=1}^K$ as keys and values to conduct multi-head attention. Moreover, for vision to text fusing, we use the text tokens $\boldsymbol{W}=\{\boldsymbol{w}_i\}_{i=1}^K$ as queries, the vision tokens $\boldsymbol{G}=\{{\boldsymbol{g}_i}\}_1^M$ as keys and values to conduct MSA. 

\section{Experiments}

\begin{table}[]
	\centering
	\begin{adjustbox}{max width=0.5\textwidth}
		\begin{tabular}{lcccc}
			\toprule[1pt]
			\multicolumn{1}{l|}{\multirow{2}{*}{Model}} & \multicolumn{2}{c|}{\textbf{VQA2.0}}                 & \multicolumn{2}{c}{\textbf{NLVR2}}          \\
			\multicolumn{1}{l|}{}                       & test-dev       & \multicolumn{1}{c|}{test-std}       & dev            & \multicolumn{1}{l}{test-p} \\ \midrule
			\multicolumn{5}{l}{\textit{Pre-trained with \textgreater{}10M images}}                                                                           \\ \midrule
			\multicolumn{1}{l|}{ALBEF(14M)~\cite{li2021albef}}             & 75.84          & \multicolumn{1}{c|}{76.04}          & 82.55          & 83.14                      \\
			\multicolumn{1}{l|}{SimVLM~\cite{DBLP:conf/iclr/WangYYDT022/simvlm}}                 & 77.87          & \multicolumn{1}{c|}{78.14}          & 81.72          & 81.77                      \\
			\multicolumn{1}{l|}{OFA~\cite{DBLP:conf/icml/WangYMLBLMZZY22/ofa}}                    & 78.0           & \multicolumn{1}{c|}{78.1}           & -              & -                          \\
			\multicolumn{1}{l|}{BLIP~\cite{DBLP:conf/icml/0001LXH22/blip}}                   & 78.25          & \multicolumn{1}{c|}{78.32}          & 82.15          & 82.24                      \\ \midrule
			\multicolumn{5}{l}{\textit{Pre-trained with \textless{}10M images}}                                                                              \\ \midrule
			\multicolumn{1}{l|}{Oscar~\cite{DBLP:conf/eccv/Li0LZHZWH0WCG20/oscar}}                  & 73.16          & \multicolumn{1}{c|}{73.44}          & 78.07          & 78.36                      \\
			\multicolumn{1}{l|}{UNITER~\cite{chen2020uniter}}                 & 72.70          & \multicolumn{1}{c|}{72.91}          & 77.18          & 77.85                      \\
			\multicolumn{1}{l|}{ViLT~\cite{DBLP:conf/icml/KimSK21/vilt}}                   & 71.26          & \multicolumn{1}{c|}{-}              & 75.70           & 76.13                      \\
			\multicolumn{1}{l|}{TCL~\cite{DBLP:conf/cvpr/YangDTXCCZCH22/tcl}}                    & 74.90          & \multicolumn{1}{c|}{74.92}          & 80.54          & 81.33                      \\
			\multicolumn{1}{l|}{VLMo\cite{DBLP:journals/corr/abs-2111-02358/vlmo}}                 & 76.64          & \multicolumn{1}{c|}{76.89}          & 82.77          & 83.34                      \\
			\multicolumn{1}{l|}{METER~\cite{Dou_2022_CVPR/meter}}                  & 77.68          & \multicolumn{1}{c|}{77.64}          & 82.33          & 83.05                      \\
			\multicolumn{1}{l|}{Ours}                   & \textbf{78.72} & \multicolumn{1}{c|}{\textbf{78.78}} & \textbf{83.63} & \textbf{84.27}             \\ \bottomrule[1pt]
		\end{tabular}
	\end{adjustbox}
	\vspace{-0.2cm}
	\caption{Performance comparison on VQA2.0 and NLVR2.}
	\label{table:vqanlvr2}
	\vspace{-0.5cm}
\end{table}

% 保留小数位
\begin{table}[]
	\centering
	\small
	\begin{adjustbox}{max width=0.5\textwidth}
		\begin{tabular}{lcccccc}
			\toprule[1pt]
			\multicolumn{1}{l|}{\multirow{2}{*}{Model}} & \multicolumn{6}{c}{\textbf{Flicker30K-ZS}}                                          \\
			\multicolumn{1}{c|}{}                       & IR@1  & IR@5  & \multicolumn{1}{l}{IR@10} & TR@1 & TR@5 & \multicolumn{1}{l}{TR@10} \\ \midrule
			\multicolumn{7}{l}{\textit{Evaluate pre-trained models directly}}                                                                 \\ \midrule
			\multicolumn{1}{l|}{UNITER~\cite{chen2020uniter}}           & 66.16 & 88.40  & 92.94                     & 80.70 & 95.70 & 98.00                     \\
			\multicolumn{1}{l|}{ViLT~\cite{DBLP:conf/icml/KimSK21/vilt}}                   & 55.0    & 82.5  & 89.8                      & 73.2 & 93.6 & 96.5                      \\
			\multicolumn{1}{l|}{ALIGN~\cite{DBLP:conf/icml/JiaYXCPPLSLD21/align}}                 & 75.70    & 93.80  & 96.80                      & 88.60 & 98.70 & 99.70                      \\
			\multicolumn{1}{l|}{METER~\cite{Dou_2022_CVPR/meter}}         & 79.60 & 94.96          & 97.28                     & 90.90          & 98.30          & 99.50                      \\
			\multicolumn{1}{l|}{Ours}                   & \textbf{79.74}         & \textbf{95.46} & \textbf{97.86}            & \textbf{91.70} & \textbf{99.30} & \textbf{99.90}        \\ \midrule
			\multicolumn{7}{l}{\textit{Evaluate models fine-tuned on COCO}}                                                                    \\ \midrule
			\multicolumn{1}{l|}{ALBEF~\cite{li2021albef}}            & 76.8  & 93.7  & 96.7                      & 90.5 & 98.8 & 99.7                      \\
			\multicolumn{1}{l|}{ALBEF\protect\footnotemark[1]~\cite{li2021albef}}           & \textbf{82.8}  & 96.3  & 98.1                      & 94.1 & 99.5 & 99.7                      \\
			\multicolumn{1}{l|}{TCL~\cite{DBLP:conf/cvpr/YangDTXCCZCH22/tcl}}                    & 79.6  & 95.1  & 97.4                      & 93.0   & 99.1 & 99.6                      \\
			\multicolumn{1}{l|}{Ours}                   & 81.74         & \textbf{96.72} & \textbf{98.54}            & \textbf{94.80} & \textbf{99.60} & \textbf{100.00}                      \\ 
			\bottomrule[1pt]
		\end{tabular}
	\end{adjustbox}
	\vspace{-0.2cm}
	\caption{Performance comparison of zero-shot image-text retrieval on Flickr30K.}
	\label{table:f30k-zs}
	\vspace{-0.3cm}
\end{table}

\footnotetext[1]{Pre-trained on 14M images.}
\subsection{Implementation Details}
Following a recent line of works, we use COCO~\cite{DBLP:conf/eccv/LinMBHPRDZ14/mscoco}, Visual Genome (VG)~\cite{DBLP:journals/ijcv/KrishnaZGJHKCKL17/vg}, Conceptual Captions (CC3M)~\cite{DBLP:conf/acl/SoricutDSG18/cc3m}, and SBU Captions~\cite{DBLP:conf/nips/OrdonezKB11/sbu} for image-text pre-training, which contain 4M images in total. 
Then the pre-trained model is applied to image-text downstream tasks and the initialization for video-text pre-training. 
For the following video-text pre-training, we utilize WebVid~\cite{bain2021frozen} with 2.5M videos as the pre-training corpus. 
We employ an extensive set of evaluation benchmarks on a wide variety of vision-language understanding and retrieval tasks, including visual question answering (VQA2.0~\cite{balanced_vqa_v2/vqa_v2}), visual reasoning (NLVR2~\cite{DBLP:conf/acl/SuhrZZZBA19/nlvr2}), image-text retrieval (Flickr30K~\cite{DBLP:conf/iccv/PlummerWCCHL15/f30k}, COCO~\cite{DBLP:conf/eccv/LinMBHPRDZ14/mscoco}), and video-text retrieval (MSRVTT~\cite{DBLP:conf/cvpr/XuMYR16/msrvtt}, LSMDC~\cite{DBLP:conf/cvpr/RohrbachRTS15/lsmdc}). 
We utilize CLIP-ViT-224/16~\cite{DBLP:conf/icml/RadfordKHRGASAM21/clip} and RoBERTa~\cite{DBLP:journals/corr/abs-1907-11692/roberta} to initialize vision and language encoders following METER~\cite{Dou_2022_CVPR/meter}. 
For our proposed SCL, we adopt high mask ratios, 80$\%$ for images and 40$\%$ for texts. 
Other details of pre-training settings can be found in~\cref{appendix:exp}. 
% Details of  can be founded in~\cref{appendix:exp}. % 最后看空间而定是否把介绍放在正文

% We utilize CLIP-ViT-224/16~\cite{DBLP:conf/icml/RadfordKHRGASAM21/clip} and RoBERTa~\cite{DBLP:journals/corr/abs-1907-11692/roberta} to initialize vision and language encoders following METER~\cite{Dou_2022_CVPR/meter}. 
% The fusion encoder consists of dual-stream cross-modal blocks of 6 layers, each with a hidden dimension of 768 and 12 heads in the multi-head attention. 
% As for data pre-processing, the image size is set to $288\times 288$ for pre-training and $384\times 384$ for fine-tuning, respectively. 
% RandAugment~\cite{DBLP:conf/nips/CubukZS020/randaug} is applied for data augmentation. 
% We resize each frame of video to $224\times 224$ and uniformly sample 4 frames as video input. % 均匀吗
% For our proposed SCL, we adopt high mask ratios, 80$\%$ for images and 40$\%$ for texts. 
% We pre-train the model for 100k steps totally using a batch size of 4096 on 64 NVIDIA A100 GPUs. 
% We adopt the AdamW optimizer with a weight decay of 0.02. % ?
% The learning rate of feature extractors is warmed up from 0 to $1e-5$ in first $10\%$ steps and then decayed linearly. 
% The fusion transformer has a five times higher learning rate. 
% The temperature hyper-parameter $\tau$ in Eq.~\ref{eq:nceloss} is set as 0.03. 

\begin{table*}[]
	\centering
	\begin{adjustbox}{max width=\textwidth}
		\begin{tabular}{lcccccccccccc}
			\toprule[1pt]
			\multicolumn{1}{l|}{\multirow{2}{*}{Model}} & \multicolumn{6}{c|}{\textbf{COCO}}                                                                                                                             & \multicolumn{6}{c}{\textbf{Flickr30K}}                                                                                                                 \\
			\multicolumn{1}{l|}{}                       & IR@1           & IR@5           & \multicolumn{1}{l}{IR@10} & \multicolumn{1}{l}{TR@1} & \multicolumn{1}{l}{TR@5} & \multicolumn{1}{l|}{TR@10}          & IR@1           & IR@5           & \multicolumn{1}{l}{IR@10} & \multicolumn{1}{l}{TR@1} & \multicolumn{1}{l}{TR@5} & \multicolumn{1}{l}{TR@10} \\ \midrule
			\multicolumn{13}{l}{\textit{Pre-trained with \textgreater{} 10M images}}                                                                                                                                                                                                                                                                                                    \\ \midrule
			\multicolumn{1}{l|}{ALIGN~\cite{DBLP:conf/icml/JiaYXCPPLSLD21/align}}                  & 59.9           & 83.3           & 89.8                      & 77.0                     & 93.5                     & \multicolumn{1}{c|}{96.9}           & 84.9           & 97.4           & 98.6                      & 95.3                     & 99.8                     & 100.0                       \\
			\multicolumn{1}{l|}{ALBEF(14M)~\cite{li2021albef}}             & 60.7           & 84.3           & 90.5                      & 77.6                     & 94.3                     & \multicolumn{1}{c|}{97.2}           & 85.6           & 97.5           & 98.9                      & 95.9                     & 99.8                     & 100.0                     \\ \midrule
			\multicolumn{13}{l}{\textit{Pre-trained with \textless{} 10M images}}                                                                                                                                                                   \\ \midrule
			\multicolumn{1}{l|}{UNITER~\cite{chen2020uniter}}                 & 52.93          & 79.93          & 87.95                     & 65.68                    & 88.56                    & \multicolumn{1}{c|}{93.76}          & 75.56          & 94.08          & 96.76                     & 87.30                    & 98.00                    & 99.20                     \\
			\multicolumn{1}{l|}{PixelBERT~\cite{DBLP:journals/corr/abs-2004-00849/pixel-bert}}              & 50.1           & 77.6           & 86.2                      & 63.6                     & 87.5                     & \multicolumn{1}{c|}{93.6}           & 71.5           & 92.1           & 95.8                      & 87.0                     & 98.9                     & 99.5                      \\
			\multicolumn{1}{l|}{VinVL~\cite{DBLP:conf/cvpr/ZhangLHY0WCG21/vinvl}}                  & 58.1           & 83.2           & 90.1                      & 74.6                     & 92.6                     & \multicolumn{1}{c|}{96.3}           & -              & -              & -                         & -                        & -                        & -                         \\
			\multicolumn{1}{l|}{ALBEF(4M)~\cite{li2021albef}}              & 56.80          & 81.50          & 89.20                     & 73.10                    & 91.40                    & \multicolumn{1}{c|}{96.00}          & 82.80          & 96.70          & 98.40                     & 94.30                    & 99.40                    & 99.80                     \\
			\multicolumn{1}{l|}{SOHO~\cite{huang2021seeing/soho}}                   & 50.6           & 78.0           & 86.7                      & 66.4                     & 88.2                     & \multicolumn{1}{c|}{93.8}           & 72.5           & 92.7           & 96.1                      & 86.5                     & 98.1                     & 99.3                      \\
			\multicolumn{1}{l|}{TCL~\cite{DBLP:conf/cvpr/YangDTXCCZCH22/tcl}}                    & 59.0           & 83.2           & 89.9                      & 75.6                     & 92.8                     & \multicolumn{1}{c|}{96.7}           & 84.0           & 96.7           & 98.5                      & 94.9                     & 99.5                     & 99.8                      \\
% 			\multicolumn{1}{l|}{VLMo~\cite{DBLP:journals/corr/abs-2111-02358/vlmo}}                   & 57.2           & 82.6           & 89.8                      & 74.8                     & 93.1                     & \multicolumn{1}{c|}{96.9}           & 79.3           & 95.7           & 97.8                      & 92.3                     & 99.4                     & 99.9                      \\
			\multicolumn{1}{l|}{METER~\cite{Dou_2022_CVPR/meter}}                  & 57.08          & 82.66          & 90.07                     & 76.16                    & 93.16                    & \multicolumn{1}{c|}{96.82}          & 82.22          & 96.34          & 98.36                     & 94.30                    & 99.60                    & 99.90                     \\
			\multicolumn{1}{l|}{Ours}                   & \textbf{60.14} & \textbf{84.56} & \textbf{91.45}            & \textbf{77.70}            & \textbf{94.10}            & \multicolumn{1}{c|}{\textbf{97.44}} & \textbf{84.56} & \textbf{97.42} & \textbf{98.94}            & \textbf{95.90}            & \textbf{99.80}            & \textbf{100.00}                     \\ \bottomrule[1pt]
		\end{tabular}
	\end{adjustbox}
	\vspace{-0.2cm}
	\caption{Performance comparison of fine-tuned image-text retrieval on Flickr30K and COCO datasets.}
	\label{table:f30k_coco}
	\vspace{-0.3cm}
\end{table*}

\begin{table}[]
	\centering
	\begin{adjustbox}{max width=0.5\textwidth}
		\begin{tabular}{lccc|ccc}
			\toprule[1pt]
			\multicolumn{1}{l|}{\multirow{2}{*}{Model}} & \multicolumn{3}{c|}{\textbf{MSRVTT}}                 & \multicolumn{3}{c}{\textbf{LSMDC}}          \\
			\multicolumn{1}{l|}{}                       & R@1    & R@5    & R@10             & R@1    & R@5    & R@10        \\ \midrule
			\multicolumn{7}{l}{\textit{Fine-tune}}                                                                           \\ \midrule
			\multicolumn{1}{l|}{VideoCLIP~\cite{xu2021videoclip}}             &30.9     &55.4.  &66.8        &-     &-     &-                  \\
			\multicolumn{1}{l|}{Frozen~\cite{bain2021frozen}}                & 31.0    & 59.5     &70.5         &15.0     &30.8     &39.8               \\
			\multicolumn{1}{l|}{VIOLET~\cite{fu2021violet}}                 &34.5    & 63.0    & 73.4      &16.1     &36.6     &41.2             \\
			\multicolumn{1}{l|}{ALPRO~\cite{DBLP:conf/cvpr/Li0LNH22/alpro}}                  &33.9    & 60.7    & 73.2      &-     &-     &-        \\ 
			\multicolumn{1}{l|}{MCQ~\cite{DBLP:conf/cvpr/GeGLLSQL22/mcq}}                  &37.6    & 64.8    & 75.1      &17.9     &35.4     &44.5             \\ 
			\multicolumn{1}{l|}{MILES~\cite{DBLP:journals/corr/abs-2204-12408/miles}}                 &37.7     &63.6   &73.8         &17.8     &35.6      &44.1                    \\
			\multicolumn{1}{l|}{Clover~\cite{DBLP:journals/corr/abs-2207-07885/clover}}                &38.6    & 67.4    & 76.4      &22.7     &42.0     &52.6            \\ \midrule
			\multicolumn{1}{l|}{Ours}                &\textbf{43.2}   & \textbf{76.0 }   & \textbf{86.7}      &\textbf{32.8}    &\textbf{62.5}    &\textbf{72.9 }       \\ \midrule
			\multicolumn{7}{l}{\textit{Zero-shot}}                                                                              \\ \midrule
			\multicolumn{1}{l|}{VideoCLIP~\cite{xu2021videoclip}}             &10.4     &22.2   &30.0            &-     &-     &-                       \\
			\multicolumn{1}{l|}{Frozen~\cite{bain2021frozen}}                 &18.7     &39.6  &51.6             &9.3     &22.0     &30.1                  \\
			\multicolumn{1}{l|}{VIOLET~\cite{fu2021violet}}                   &25.9     &49.5   &59.7            &-     &-     &-               \\
			\multicolumn{1}{l|}{ALPRO~\cite{DBLP:conf/cvpr/Li0LNH22/alpro}}                   &24.1     &44.7  &55.4       &-     &-     &-                       \\ 
			\multicolumn{1}{l|}{MCQ~\cite{DBLP:conf/cvpr/GeGLLSQL22/mcq}}                   &26.0     &46.4   &56.4         &12.2      &25.9      &32.2                     \\ 
			\multicolumn{1}{l|}{MILES~\cite{DBLP:journals/corr/abs-2204-12408/miles}}                 &26.1     &47.2   &56.9         &11.1     &24.7      &30.6                    \\
			\multicolumn{1}{l|}{Clover~\cite{DBLP:journals/corr/abs-2207-07885/clover}}                 &25.8     &49.6   &60.1         & 13.8      &28.1      &38.3         \\         \midrule
			\multicolumn{1}{l|}{Ours}                &\textbf{30.9}   & \textbf{54.4}    & \textbf{65.0}     &\textbf{17.2}    &\textbf{32.4}    &\textbf{39.1}       \\
			\bottomrule[1pt]
		\end{tabular}
	\end{adjustbox}
	\vspace{-0.2cm}
	\caption{Performance comparison of text-to-video retrieval on MSRVTT and LSMDC.}
	\label{table:msrvtt_lsmdc}
	\vspace{-0.3cm}
\end{table}

\subsection{Evaluation Results}

\subsubsection{Image-Text Understanding}
We conduct multimodal understanding tasks on VQA2.0 and NLVR2, which require the model to exploit vision and language semantic fusion. 
As the results shown in Table~\ref{table:vqanlvr2}, SCL achieves new state-of-the-art performance compared with previous models, implying that cross-modal fusion benefits from our new pre-training task. 
Specifically, when pre-trained with fewer than 10M images, our model outperforms METER~\cite{Dou_2022_CVPR/meter} by $+1.04$ and $+1.14$ scores on VQA2.0 test-dev and test-std. 
On NLVR2, we gain $+0.86$ and $+0.93$ score improvements over VLMo~\cite{DBLP:journals/corr/abs-2111-02358/vlmo}, respectively. 
Moreover, our model pre-trained with 4M images also surpasses some models with more than 10M images, for instance, SimVLM~\cite{DBLP:conf/iclr/WangYYDT022/simvlm}, BLIP~\cite{DBLP:conf/icml/0001LXH22/blip}.

\subsubsection{Image-Text Retrieval}
We evaluate image-text retrieval in both zero-shot and fine-tuning scenarios. 
Our model achieves substantial performance improvements on Flickr30K and COCO datasets with similar model sizes and pre-training data scales. 
The results are also competitive with models pre-trained on larger datasets, such as  ALIGN~\cite{DBLP:conf/icml/JiaYXCPPLSLD21/align} and ALBEF~\cite{li2021albef}. 
In the fine-tuning phase, the model is trained with CL and VTM losses. 
During inference, for the sake of efficiency, we first filter top-k candidates with vision and language encoders and then compute VTM scores for ranking. 

%We leverage zero-shot experiments to demonstrate our model's generalization ability. 
To investigate the generalization ability of our model, we conduct zero-shot experiments on the Flickr30K dataset. 
As shown in Table~\ref{table:f30k-zs}, SCL achieves the best performance in both settings of zero-shot retrieval on Flickr30K. 
%, which is evaluated with the pre-trained model directly or the model fine-tuned on COCO. % 模型名字
When we evaluate with the pre-trained model directly, SCL gains a comprehensive boost from previous methods, reaching 79.74$\%$ and 91.7$\%$ in terms of IR@1 and TR@1. 
When evaluated with the model fine-tuned on COCO, SCL outperforms models pre-trained on datasets of similar sizes, including ALBEF~\cite{li2021albef} and TCL~\cite{DBLP:conf/cvpr/YangDTXCCZCH22/tcl}.  
Moreover, compared with ALBEF pre-trained on 14M images, SCL also has a more impressive performance in five out of six recall metrics, further demonstrating the effectiveness of our proposed strategies. 

For fine-tuning experiments, our model surpasses previous models by a large margin, as shown in Table~\ref{table:f30k_coco}. 
TCL~\cite{DBLP:conf/cvpr/YangDTXCCZCH22/tcl} has a distinguished retrieval performance with triple contrastive learning, which is cross-modal, intra-modal, and global-local. 
Compared with TCL, our method brings $+1.14\%/+0.56\%$ IR@1 boost and $+2.10\%/+1.00\%$ TR@1 on COCO and Flickr30K, respectively. 
It is worth noting that our model also has higher scores than ALIGN~\cite{DBLP:conf/icml/JiaYXCPPLSLD21/align} with 1.8B image-text pairs pre-trained. 
Thanks to semantic completion learning, the global features capture more cross-modal information, leading to an encouraging performance on retrieval. 

\subsubsection{Video-Text Retrieval}
Due to our adaptable vision encoder, the image-text pre-trained model can be readily transferred to video-text pre-training. 
We evaluate text-to-video retrieval on two popular datasets, MSRVTT and LSMDC, to prove the performance of the video pre-training model. 
Table~\ref{table:msrvtt_lsmdc} summarizes results under both fine-tuning and zero-shot settings. 
In the fine-tuning situation, compared with the previous SOTA model, SCL achieves notable performance improvements with $+4.6\%$ and $+10.1\%$ in R@1 on MSRVTT and LSMDC. 
When doing zero-shot retrieval, SCL also gains remarkable improvements over the existing methods with $+4.8\%$ and $+3.4\%$ in R@1 on MSRVTT and LSMDC, respectively. 
These results demonstrate that the knowledge of our SCL model learned from image-text data can be used to improve the performance of video-text retrieval tasks.

\begin{table*}[t]
	\centering
	\begin{adjustbox}{max width=\textwidth}
		\begin{tabular}{c|cccccc|c|cc}
			\toprule[1pt]
			\multirow{2}{*}{Pre-training tasks} & \multicolumn{6}{c|}{\textbf{Flicker30K-ZS}}          & \textbf{VQA2.0}   & \multicolumn{2}{c}{\textbf{NLVR2}} \\
			& IR@1  & IR@5  & IR@10 & TR@1 & TR@5 & TR@10 & test-dev & dev         & test-p      \\ \midrule
			Ours (MLM+VTM+CL+SCL)            & \textbf{70.4} & \textbf{91.1} & \textbf{95.2} & \textbf{86.2} & \textbf{97.3} & 99.2  & 77.59    & 81.01       & 81.89       \\
			w/o MLM                             & 67.6 & 89.5 & 94.4 & 84.7 & 97.5 & 99.3  & 75.46    & 78.93       & 79.41       \\
			w/o VTM                             &   49.7    &   78.5   &  87.2    &  61.1   &  86.4  &    94.7  & 76.96    & 77.86       & 79.54       \\
			w/o CL                              & 67.8 & 90.7 & 94.9 & 82.5 & 97.3 & 99.1  & \textbf{77.66}    & \textbf{81.10}        & \textbf{82.18}       \\   % \hdashline 
			w/o SCL                             & 67.0 & 89.4 & 94.3 & 80.0   & 96.5 & \textbf{99.4}  & 77.31    & 80.30        & 81.46      \\
			\bottomrule[1pt]
		\end{tabular}
	\end{adjustbox}
	\vspace{-0.2cm}
	\caption{Ablation study of each pre-training task. Note that the model pre-trained without VTM can only conduct zero-shot retrieval with vision and text encoders without feature fusion, so the recall metrics are not comparable to the others.}
	\label{table:pre-training tasks}
	\vspace{-0.1cm}
\end{table*}

\begin{table}[]
	\begin{adjustbox}{max width=0.5\textwidth}
		\begin{tabular}{c|cccc|c|c}
			\toprule[1pt]
			\multirow{2}{*}{Pre-training tasks} & \multicolumn{4}{c|}{\textbf{Flickr30K-ZS}}                          & \textbf{VQA2.0}   & \textbf{NLVR2}  \\
			& IR@1                      & IR@5  & \multicolumn{1}{l}{TR@1} & TR@5 & test-dev & test-p \\ \midrule
			MLM+VTM+CL                    & 67.0                     & 89.4 & 80.0                       & 96.5 & 77.31    & 81.46  \\
			+MLSC                               & 67.0                     & 90.0 & 80.5                    & 97.1 & \textbf{77.60}     & 81.88  \\
			+MVSC                               & 69.5                    & 90.5 & 85.5                     & \textbf{98.3} & 77.50     & 81.69  \\
			+SCL         & \textbf{70.4} & \textbf{91.1} & \textbf{86.2}  & 97.3 & 77.59    & \textbf{81.89}  \\ \bottomrule[1pt]
		\end{tabular}
	\end{adjustbox}
	\vspace{-0.2cm}
	\caption{Ablation study of MLSC and MVSC.}
	\label{table:scl pre-training tasks}
	\vspace{-0.2cm}
\end{table}

\subsection{Ablation Studies}

We conduct empirical ablation experiments on pre-training tasks and the vision encoder. 
Since pre-training is time-consuming, we use COCO and VG as pre-training datasets, which is also a common setting in previous works~\cite{DBLP:conf/emnlp/TanB19/lxmert, DBLP:conf/acl/XuYLBHXH20/e2e-vlp, huang2021seeing/soho}. 

\subsubsection{Different Pre-training Tasks}

\begin{table}[]
	\begin{adjustbox}{max width=0.5\textwidth}
		\begin{tabular}{cl|cccc|c}
			\toprule[1pt]
			\multicolumn{2}{c|}{mask ratio} & \multicolumn{4}{c|}{\textbf{Flicker30K-ZS}}                              & \textbf{VQA2.0}         \\
			image          & text           & IR@1           & IR@5           & TR@1          & TR@5          & test-dev       \\ \midrule
			0.7            & 0.4            & 68.8          & 90.6         & 84.0            & \textbf{97.7} & 77.51          \\
			0.8            & 0.3            & 69.5          & 90.5         & 83.8          & 97.6          & \textbf{77.63} \\
			0.8   & 0.4   & \textbf{70.4} & \textbf{91.1} & \textbf{86.2} & 97.3          & 77.59          \\
			0.8            & 0.5            & 68.8           & 90.5         & 81.9           & 97.4           & \textbf{77.63} \\
			0.9            & 0.4            & 68.6           & 90.6           & 82.8          & 97.3           & 77.57          \\ 
			\bottomrule[1pt]
		\end{tabular}
	\end{adjustbox}
	\vspace{-0.2cm}
	\caption{Effect of different image and text mask ratios.}
	\label{table：mask ratio}
	\vspace{-0.2cm}
\end{table}

\begin{table}[]
	\centering
	\begin{adjustbox}{max width=0.5\textwidth}
		\begin{tabular}{lccc|ccc}
			\toprule[1pt]
			\multicolumn{1}{l|}{\multirow{2}{*}{Vision Encoder}} & \multicolumn{3}{c|}{\textbf{Video Retrieval}}                 & \multicolumn{3}{c}{\textbf{Text Retrieval}}          \\
			\multicolumn{1}{l|}{}                       & R@1    & R@5    & R@10             & R@1    & R@5    & R@10        \\ \midrule
			\multicolumn{1}{c|}{Mean Pooling}             &23.0      &45.6      & 56.5    &22.5   & 46.4   & 54.9                 \\
			\multicolumn{1}{c|}{Global CLS}                & 22.5    & 45.9    &  55.1         &22.4   & 45.7   & 55.5            \\
			\multicolumn{1}{c|}{Frame CLS}                 &\textbf{23.3}    & \textbf{48.3}    & \textbf{56.8}     &\textbf{24.0}   &\textbf{46.5}   &\textbf{55.7}    \\
			\bottomrule[1pt]
		\end{tabular}
	\end{adjustbox}
	\vspace{-0.2cm}
	\caption{Ablation study of the vision encoder on zero-shot retrieval of MSRVTT.}
	\label{table:video_ablation}
	\vspace{-0.4cm}
\end{table}
There are four pretext tasks in our method, including CL, VTM, MLM, and SCL. 
As summarized in Table~\ref{table:pre-training tasks}, we explore the impact of each task on both retrieval and understanding datasets. 
The first row shows the results of our model with all pre-training tasks, and the second to fifth rows reflect the effect of removing each task separately. 
According to the chart, we observe that the retrieval performance drops most due to the lack of SCL when conducting retrieval with feature fusion. 
Specifically, SCL brings $+3.38\%$ and $+6.20\%$ boost in IR@1 and TR@1 on F30K-ZS. 
The model without MLM loses $2.13\%$ in the accuracy of VQA2.0, which indicates that MLM has a great effect on multimodal understanding tasks. 
As for NLVR2, VTM has a relatively large impact. 
However, contrastive learning is only effective for retrieval in our model, which is perhaps because the other three pre-training tasks have already learned cross-modal fusion sufficiently for understanding tasks. 
Overall, comparing the first row with the fifth row, the model with SCL makes progress on all downstream tasks, which demonstrates that the model learns more accurate cross-modal alignment to generate representative global features. 

Furthermore, SCL comprises MVSC and MLSC, whose effects we showcase in Table~\ref{table:scl pre-training tasks}. 
According to the first three rows, either MVSC or MLSC can improve the performance of downstream tasks. 
We find that MVSC has a superior impact on retrieval tasks, which is probably because it improves the robustness of visual information understanding. 
In VQA2.0 and NLVR2, MLSC plays a more important role. 
% The possible reason is that text of high-level
Additionally, when combining the two sub-tasks, our model performs better in most metrics, which indicates that they are in synergy. 

\subsubsection{Mask Ratio in SCL}

As shown in Table~\ref{table：mask ratio}, we observe that the mask ratios of image and text affect downstream tasks, especially on zero-shot retrieval. 
VQA2.0 is less sensitive to the mask ratio because the model has been fine-tuned with a large amount of data. 
Considering the second to fourth rows, when the image mask ratio is fixed, the model with a text mask ratio of 0.4 has almost the best performance. 
Moreover, when the text mask ratio is set to 0.4, the results of the image mask ratio of 0.8 are the highest. 
We speculate that when the mask ratio is lower, semantic completion will rely more on intra-modal information and lack learning across modalities, leading to inferior performance. 
When the mask ratio is too high, the small number of remaining tokens can only perform very limited cross-modal interactions. 
In conclusion, we choose 0.4 and 0.8 as text and image mask ratios, respectively. 
% according to comprehensive results on F30K-ZS and VQA2.0. 

\begin{figure*}[tp]
	\centering
	\includegraphics[width=\textwidth]{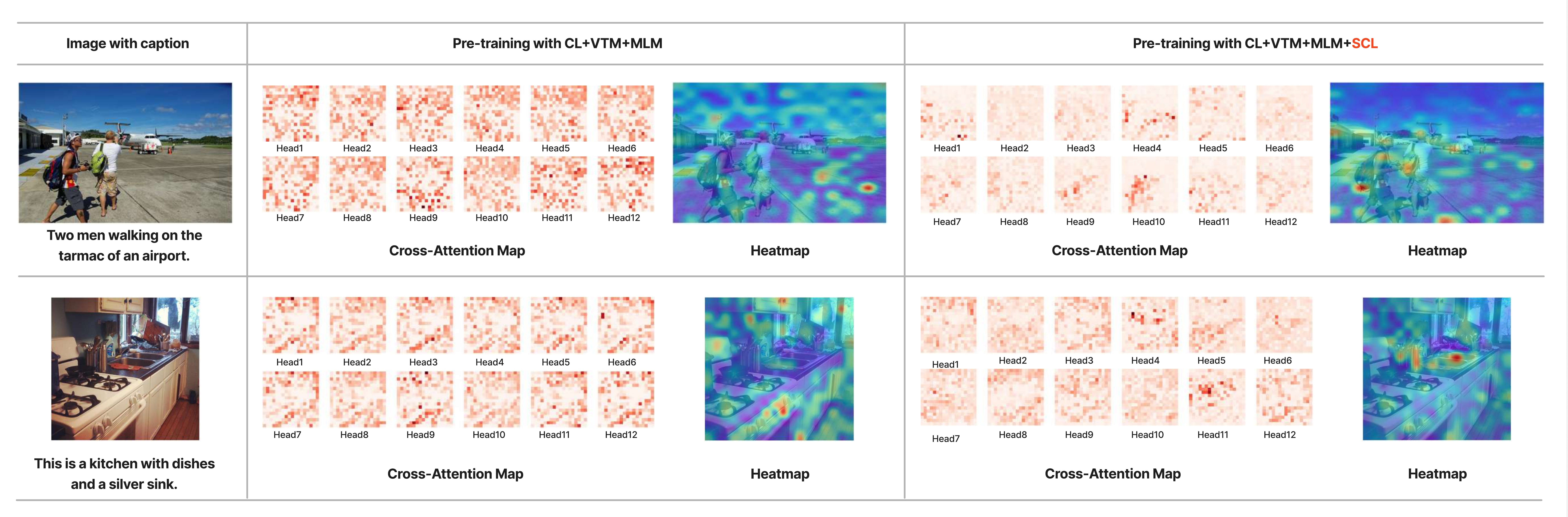}
	\caption{The cross-attention visualization of text [CLS] on the whole image for the model pre-trained with or without SCL. The cross-modal attention maps of 12 heads are from the last layer of the fusion encoder. Then we depict heatmaps by max-pooling the attention maps.}
	\label{fig:visualization}
	\vspace{-0.2cm}
\end{figure*}

\subsubsection{Vision Encoder Design}
To investigate the effectiveness of our designed vision encoder in processing video data, we compare it with two other variants: (1) The first variant, termed Mean Pooling, directly treats a video as $M$ separate images and then uses the mean pooling of $M$ [CLS] tokens as the video representation. (2) The second variant, termed Global CLS, is the vision encoder proposed by MCQ~\cite{DBLP:conf/cvpr/GeGLLSQL22/mcq}. 
In this experiment, we pre-train the vision encoder and text encoder on the WebVid~\cite{bain2021frozen} dataset via contrastive learning and then conduct zero-shot cross-modal retrieval on the MSRVTT dataset. The experimental results are shown in Table \ref{table:video_ablation}, where the Frame CLS denotes our designed vision encoder. It can be found that Frame CLS achieves the best performance on both video-to-text and text-to-video retrieval tasks, which demonstrates the outstanding capability of video temporal modeling.

\subsection{Visualization Analysis}
\label{sec:visualization}
To demonstrate that SCL boosts cross-modal alignment for global representations, we visualize cross-attention maps between text [CLS] and the whole image in the last layer of the fusion encoder. 
We conduct max-pooling on attention maps of 12 heads to draw heatmaps, as shown in Fig.~\ref{fig:intro_visulize}(b) and Fig.~\ref{fig:visualization}. 
Compared with pre-trained by CL, VTM and MLM, the model with SCL can recognize relevant regions more precisely. 
For example, in the first image of Fig~\ref{fig:intro_visulize}(b), the attention distribution of global text representation to the image is scattered without SCL, while after semantic completion learning, [CLS] pays attention to the fish, lemons, asparagus in the image. 
Taking the second image of Fig~\ref{fig:visualization} as another example, the model pre-trained with SCL identifies the dishes and sink in the kitchen, which indicates a desirable global-to-local alignment ability. 

Observing attention maps of 12 heads in Fig.~\ref{fig:visualization}, we find that the attention maps without SCL are basically the same for an image, but for the model pre-trained with SCL, the attention maps of different heads are distinctive, which means that each head learns various information from the image. 
Overall, semantic completion learning encourages global representations to learn cross-modal interactions, extracting useful knowledge from the other modality. 
There are more visualization cases in~\cref{appendix:visual}. 

\section{Conclusion}
In this paper, we proposed a new vision-language pre-training task called Semantic Completion Learning (SCL). 
Different from previous pre-training tasks that reconstruct masked local tokens, SCL leverages cross-modal interactions to recover global semantic information of masked data, promoting cross-modal alignment for global representations. 
Ablation studies and visualization analysis demonstrate the effectiveness of SCL. 
Moreover, we introduced a flexible vision encoder, which adapts to image-text and video-text multimodal tasks readily. 
We conducted image-text and video-text pre-training sequentially and applied our model to various challenging downstream tasks. 
The extensive evaluations validate the great superiority of our SCL method. 

\section{Acknowledgements}
This work was partly supported by the National Natural Science Foundation of China (Grant No.  61991450) and the Shenzhen Science and Technology Program (JSGG20220831093004008; ZDSYS20200811142605016).

%%%%%%%%% REFERENCES
{\small
\bibliographystyle{ieee_fullname}
\bibliography{egbib}
}
\clearpage
\appendix
\label{sec:appendix}

\section*{Supplementary}

\section{Model Details}
\label{appendix:model}

\begin{figure*}[b]
	\centering
	\includegraphics[width=0.8\textwidth]{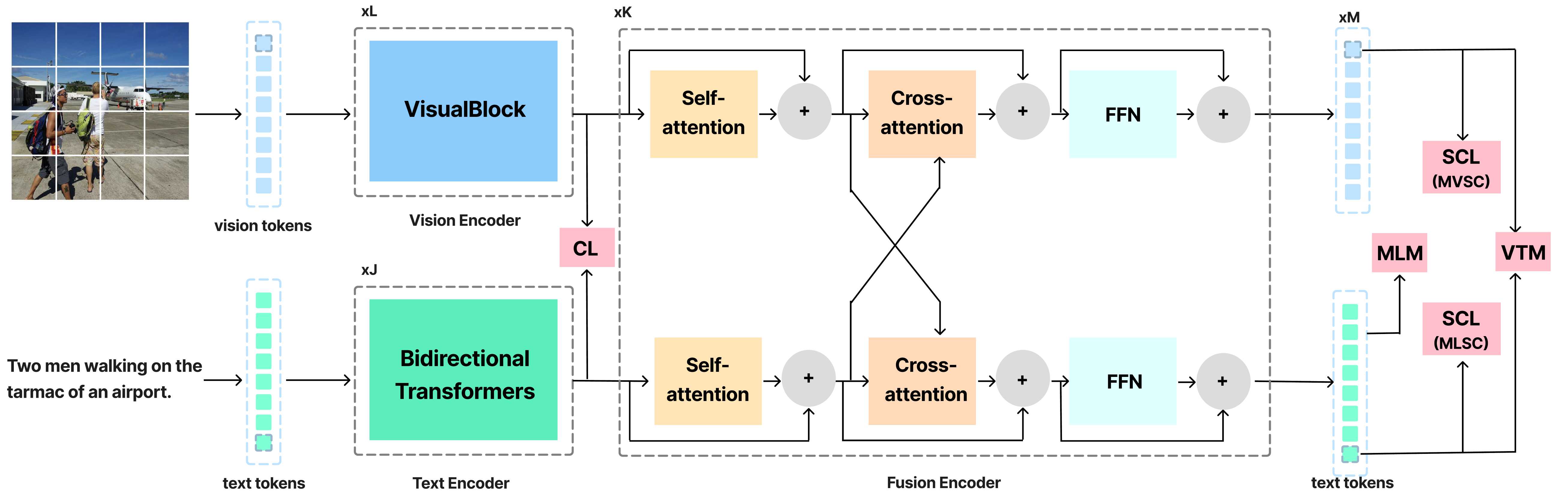}
	\vspace{-0.2cm}
	\caption{The whole architecture of our model.}
	\label{fig:architecture}
	\vspace{-0.2cm}
\end{figure*}
\subsection{Model Architecture}
% \vspace{-0.2cm}
Our model utilizes the vision encoder and text encoder to learn uni-modal representations, and the fusion encoder to conduct cross-modal interactions, respectively. 
% The pre-training tasks includes contrastive learning (CL), vision text matching (VTM), masked language modeling (MLM) and semantic completion learning (SCL). 
The whole architecture is displayed in Fig.~\ref{fig:architecture}.

\subsection{Previous Pre-training Tasks}
% \vspace{-0.2cm}
% In the following, we will introduce previous pre-training tasks in details. 

\noindent\textbf{Contrastive Learning (CL).} 
We conduct CL on the global representations from vision and text encoders. 
Given a batch of image-text pairs, for an image (text), the paired text (image) is tread as the positive sample, and other texts (images) are negative samples. 
We use the InfoNCE loss as follows:
% \vspace{-0.2cm}
\begin{equation}
\begin{aligned}
\operatorname{\textbf{NCE}}_{V2T}=-\frac{1}{N}\sum_{i=1}^{N}\operatorname{log}\frac{\operatorname{exp}(s(V_i,T_i)/\tau)}{\sum_{n=1}^N\operatorname{exp}(s(V_i,T_n)/\tau)}\, , \\
\operatorname{\textbf{NCE}}_{T2V}=-\frac{1}{N}\sum_{i=1}^{N}\operatorname{log}\frac{\operatorname{exp}(s(T_i,V_i)/\tau)}{\sum_{n=1}^N\operatorname{exp}(s(T_i,V_n)/\tau)}\, ,
\end{aligned}
\label{eq:cl nce loss}
\end{equation}
where $N$ is the batchsize and $\tau$ serves as a learnable temperature parameter. 
The similarity function is formatted as cosine similarity, $s(V,T)=\frac{\phi_v(V)^T\phi_t(T)}{||\phi_v(V)||\cdot ||\phi_t(T)||}$, where $\phi$ is a linear projection head. 
The vision-text contrastive loss is defined as:
\begin{equation}
\mathcal{L}_{CL} = \operatorname{\textbf{NCE}}_{V2T} +\operatorname{\textbf{NCE}}_{T2V}.
\end{equation}

For the video-text data, we use the mean pooling of $M$ frame [CLS] features to denote the global representation of a video and then also use Eq. (\ref{eq:cl nce loss}) for contrastive learning.

\noindent\textbf{Vision-Text Matching (VTM).} 
The model is required to predict whether a pair of image-text (video-text) is matched or not. 
Specifically, we conduct a binary classification on the concatenation of the visual and textual global features. 
The loss is defined as:
\begin{equation}
\mathcal{L}_{VTM}=\operatorname{\textbf{CE}}(\phi(\operatorname{concat}[V, T]),y),
\end{equation}
where $V$, $T$ are [CLS] features and $y$ is the ground truth. 
$\operatorname{\textbf{CE}}$ is Cross-Entropy loss and $\phi$ refers to a binary classifier. 
The image-text (video-text) pairs serve as positive samples, and we randomly replace the image (video) in a data pair with another image (video) to build negative sample. 

\noindent\textbf{Masked Language Modeling (MLM).} 
We adopt MLM following BERT~\cite{kenton2019bert}, which conducts a classification on the vocabulary list to predict masked words. 
We randomly mask out 15$\%$ text tokens, and replace them with the [MASK] token, random
words, or left unchanged, with the probability of 80$\%$, 10$\%$ and 10$\%$, respectively. 
The classification loss is as follows:
\begin{equation}
\mathcal{L}_{MLM} = \operatorname{\textbf{CE}}(\phi(T_{mask}),y),    
\end{equation}
where $T_{mask}$ is the output masked token feature, $\phi$ serves as a classifier, and $y$ is the original token ID.

The overall training objective of our model is:
\begin{equation}
\mathcal{L}=\mathcal{L}_{CL}+\mathcal{L}_{ITM}+\mathcal{L}_{MLM}+\mathcal{L}_{SCL},
\label{eq:all loss}
\end{equation}
where $\mathcal{L}_{SCL}$ is defined in Eq.~(\ref{eq:scl}).

\section{Experiments Details}
\label{appendix:exp}

\subsection{Pre-training Settings}
\vspace{-0.2cm}
In the image-text pre-training phase, we train the model for 100k steps totally using a batch size of 4096 on 64 NVIDIA A100 GPUs. 
We adopt the AdamW optimizer with a weight decay of 0.01. % ?
The learning rate of uni-modal encoders is warmed up from 0 to $1e-5$ in first $10\%$ steps and then decayed linearly. 
The fusion transformer has a five times higher learning rate. 
As for the video-text pre-training, the model is trained for 10k steps with the same batch size. 
The maximal learning rate of uni-modal encoders is $5e-6$, and other settings are similar to the first phase. 

In terms of the model architecture, We utilize CLIP-ViT-224/16~\cite{DBLP:conf/icml/RadfordKHRGASAM21/clip} and RoBERTa~\cite{DBLP:journals/corr/abs-1907-11692/roberta} to initialize vision and language encoders following METER~\cite{Dou_2022_CVPR/meter}. 
The fusion encoder consists of dual-stream cross-modal blocks of 6 layers, each with a hidden dimension of 768 and 12 heads in the multi-head attention. 
As for data pre-processing, the image size is set to $288\times 288$ for pre-training and $384\times 384$ for fine-tuning, respectively. 
RandAugment~\cite{DBLP:conf/nips/CubukZS020/randaug} is applied for data augmentation. 
We resize each frame of video to $224\times 224$ and uniformly sample 4 frames as video input. 
Moreover, the maximum length of input text is 50. 
The temperature hyper-parameter $\tau$ in Eq.~(\ref{eq:nceloss}) is set as 0.03. 

\begin{figure*}[b]
	\centering
	\includegraphics[width=0.9\textwidth]{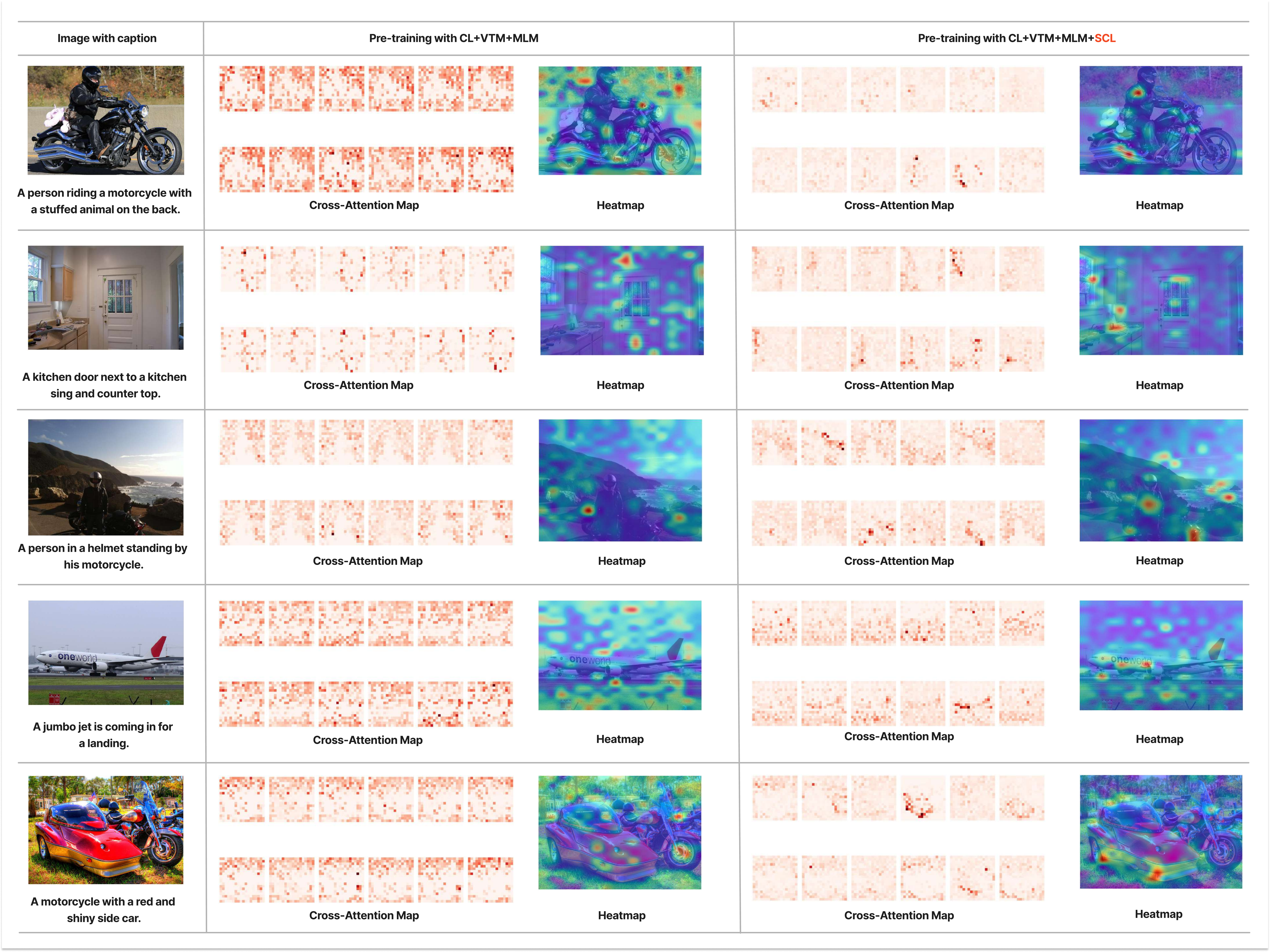}
	\caption{The cross-attention visualization of text [CLS] on the whole image for the model pre-trained with or without SCL.}
	\label{fig:appendix_visual}
	\vspace{-0.2cm}
\end{figure*}

\subsection{Downstream Tasks}
% \vspace{-0.2cm}
\noindent\textbf{Visual Question Answering (VQA).} 
Given an image and its corresponding question, the model needs to understand visual and textual information simultaneously to predict the answer. 
We concatenate output [CLS] features of the image and question, and then conduct a classification on the candidates set of 3,129 answers. 

\noindent\textbf{Visual Reasoning (NLVR2).} 
Given a pair of images and a description, the model is expected to reason whether their relationship is consistent. 
Specifically, this task is transformed to a binary classification problem. 

\noindent\textbf{Image-Text Retrieval.} 
There are two sub-tasks: (1) using images as queries to retrieve texts (TR); (2) using texts as queries to retrieve images (IR). 
The recall ratio is employed as the evaluation metrics. 
We evaluate our model on Flickr30K~\cite{DBLP:conf/iccv/PlummerWCCHL15/f30k} and COCO~\cite{DBLP:conf/eccv/LinMBHPRDZ14/mscoco}. 
Flickr30K contains 1K images and 5K texts for evaluation, and COCO includes 5K images and 25K texts. 
Generally, there are five correct captions for an image. 

\noindent\textbf{Video-Text Retrieval.} 
Similar to exiting methods~\cite{DBLP:conf/cvpr/GeGLLSQL22/mcq, fu2021violet, DBLP:journals/corr/abs-2207-07885/clover}, we focus on text-to-video recall metrics. 
Our pre-trained model is evaluated on MSRVTT~\cite{DBLP:conf/cvpr/XuMYR16/msrvtt} and LSDMC~\cite{DBLP:conf/cvpr/RohrbachRTS15/lsmdc}, which both contain 1K video-text pairs for testing. 
% \vspace{-0.2cm}
\section{Visualization Cases}
\vspace{-0.2cm}
\label{appendix:visual}
The proposed SCL encourages the global representations to learn global-to-local alignment, which implies that they have a more accurate attention distribution on local information of the other modality. 
To illustrate this, we show more visualization cases of [CLS] tokens' attention maps in Fig.~\ref{fig:appendix_visual}. 
% The depiction rule is the same as Sec.~\ref{sec:visualization}. % 两个模态的都画？

% \vspace{-0.2cm}
\section{Broader Impacts}
\vspace{-0.2cm}
Since our model predicts content based on learned statistics of pre-training datasets and we do not filter out possible inappropriate image- or video-text pairs (e.g., of violence and blood), our model may be used to retrieve unhealthy videos for spreading.

\end{document}